\documentclass[runningheads]{llncs}
\usepackage{amsmath}
\usepackage[T1]{fontenc}

\usepackage{enumitem}
\usepackage{url}
\usepackage{graphicx}
\begin{document}

\title{iText2KG: Incremental Knowledge Graphs Construction Using Large Language Models}
\titlerunning{iText2KG: Incremental Knowledge Graphs Construction Using LLMs}
%

\author{Yassir LAIRGI\inst{1,2}\orcidID{0000-0002-7284-5489} \and
Ludovic MONCLA\inst{1}\orcidID{0000-0002-1590-9546} \and
Rémy CAZABET\inst{1}\orcidID{0000-0002-9429-3865} \and
Khalid BENABDESLEM\inst{1} \and
Pierre CLÉAU\inst{2}}

\authorrunning{Y. Lairgi et al.}

%

\institute{INSA Lyon, CNRS, Universite Claude Bernard Lyon 1, LIRIS, UMR5205, 69621 Villeurbanne \email{\{ludovic.moncla, remy.cazabet, khalid.benabdeslem\}@liris.cnrs.fr} \and 
GAUC, Lyon France \email{\{yassir.lairgi, pierre.cleau\}@auvalie.com}}
\maketitle              
\begin{abstract}

Most available data is unstructured, making it challenging to access valuable information. Automatically building Knowledge Graphs (KGs) is crucial for structuring data and making it accessible, allowing users to search for information effectively. KGs also facilitate insights, inference, and reasoning.
Traditional NLP methods, such as named entity recognition and relation extraction, are key in information retrieval but face limitations, including the use of predefined entity types and the need for supervised learning.
Current research leverages large language models' capabilities, such as zero- or few-shot learning. However, unresolved and semantically duplicated entities and relations still pose challenges, leading to inconsistent graphs and requiring extensive post-processing. Additionally, most approaches are topic-dependent.
In this paper, we propose \texttt{iText2KG}\footnote{The code and the dataset are available at \url{https://github.com/AuvaLab/itext2kg}}, a method for incremental, topic-independent KG construction without post-processing. This plug-and-play, zero-shot method is applicable across a wide range of KG construction scenarios and comprises four modules: Document Distiller, Incremental Entity Extractor, Incremental Relation Extractor, and Graph Integrator and Visualization. Our method demonstrates superior performance compared to baseline methods across three scenarios: converting scientific papers to graphs, websites to graphs, and CVs to graphs.

\keywords{Knowledge Graph Construction  \and Large Language Models \and Natural Language Processing.}
\end{abstract}
\section{Introduction}

In the contemporary era, most data is unstructured, leading to substantial information loss if not effectively harnessed \cite{eberendu2016unstructured}. This unstructured data lacks a predefined format, posing significant challenges for traditional data processing methodologies. Consequently, organizations must employ advanced text understanding and information extraction techniques to analyze and extract meaningful insights from this data effectively.

Text understanding and information extraction are key tasks in Natural Language Processing (NLP) for automatically processing data from unstructured text documents.
The rise of Transformer architectures and pre-trained large language models (LLMs) opens new perspectives for extracting and structuring information from vast amounts of natural language texts \cite{jin2023large}.
One main aspect deals with Knowledge graphs (KGs) construction. KGs structure representations of knowledge by capturing relationships between entities and hold considerable advantages in analyzing text data collections and inferring knowledge from structured heterogeneous data. For instance, KGs can merge diverse data from multiple sources, offering a cohesive information perspective. They can also give an additional level of explainability to the analysis of text corpora.

Named Entity Recognition, Relation Extraction, and Entity Resolution are NLP techniques usually utilized to transform unstructured text into structured data, capturing entities, their connections, and associated attributes \cite{nasar2021named,singh2018natural}.  However, these methods encounter several limitations \cite{carta2023iterative,mihindukulasooriya2023text2kgbench}. They are frequently restricted to predefined entities and relationships or depend on specific ontologies and mostly rely on supervised learning methods, necessitating extensive human annotation.

To address these challenges, we aim to leverage LLMs in constructing KGs. Recent advancements in LLMs have shown potential and improved performance across a various range of NLP tasks, including knowledge graph completion, ontology refinement, and question answering, offering promising prospects for KG construction \cite{mihindukulasooriya2023text2kgbench}.
LLMs also show great ability for few-shot learning, enabling plug-and-play solutions, and eliminating the necessity for extensive training or fine-tuning. They can be used to extract knowledge across diverse domains due to their training in a wide range of information sources \cite{zhu2023llms}. 

Consequently, recent research has started utilizing advancements in LLMs, especially their capabilities in few-shot learning in KGs construction tasks. However, unresolved and semantically duplicated entities and relations still pose significant challenges, leading to inconsistent graphs that require extensive post-processing. These inconsistencies can manifest as redundancies, ambiguities, and a real difficulty for graph extension. Additionally, many current approaches are topic-dependent, meaning their effectiveness heavily relies on the specific use case they are designed to handle. This dependency limits the generalizability of these methods across different domains, necessitating customized solutions for each new topic area.

In this paper, we propose \texttt{iText2KG}, a zero-shot method to construct consistent KGs from raw documents incrementally, using an LLM. It comprises four modules: 1) Document Distiller reformulates the raw documents, by taking a schema or a blueprint, into predefined and semantic blocks using LLMs. The schema operates like a predefined JSON structure, directing the language model to extract specific textual information associated with particular keys from each document, 2) iEntities Extractor takes the semantic blocks and not only identifies unique semantic entities within the semantic blocks but also resolves any ambiguities, ensuring that each entity is clearly defined and distinguished from others, 3) iRelation Extractor processes the resolved entities along with the semantic blocks to detect the semantically unique relationships.  Further details are in the next sections. The final module employs Neo4j  \footnote{\url{https://neo4j.com/}} to represent these relationships and entities in a graph format visually.

\section{Related works}
\label{related_works}

LLM-based solutions for building KGs can be categorized according to three paradigms: ontology-guided, fine-tuning, and zero- or few-shot learning.

The AttacKG+ method, a fully automatic LLM-based framework for constructing attack KGs and capturing the progressive stages of cyber attacks, was introduced by \cite{zhang2024attackg+}.
The framework consists of four modules: rewriter, parser, identifier, and summarizer. The rewriter filters out redundant information and organizes report content into sections to preserve key knowledge, pre-cleans data, and sequences events chronologically. Guided by an ontology, the parser extracts threat actions using a triplet model (subject, action, object). The identifier matches these behavior graphs and rewritten sections to the appropriate format. Finally, the summarizer provides an overview of the situation and state at the end of each tactical stage. A theme-specific KG (ThemeKG) was proposed \cite{ding2024automated},  constructed from a theme-specific corpus using an unsupervised framework (TKGCon) to address two main issues: limited information granularity and deficiency in timeliness. This approach generates KGs with accurate entities and relations by leveraging common sense knowledge from Wikipedia and LLMs for ontology guidance. Their model surpasses GPT-4 in performance due to its consistently precise identification of entities and relations

Text2KGBench, a benchmark designed to evaluate the capabilities of language models to generate KGs from natural language text guided by an ontology, was presented by \cite{mihindukulasooriya2023text2kgbench}. They define seven evaluation metrics to measure fact extraction performance, ontology conformance, and hallucinations. A semi-automatic method for constructing KGs using open-source LLMs was introduced in recent research \cite{kommineni2024human}. Their pipeline includes formulating competency questions (CQs) and developing an ontology derived from them. To assess the accuracy of the generated answers, they devised a judge LLM, which evaluates the content against ground truth. One major challenge with these proposed methods is their difficulty in generalizing their applicability to diverse KG construction scenarios due to their ontology dependency. The Wikipedia concept graph is also not exhaustive, particularly for country-specific concepts. For instance, it may not adequately cover terms like "French Research Collaboration Tax Credit".

An LLM was employed for building a KG from unstructured open-source threat intelligence \cite{hu2023llm}. This approach involves generating a dataset utilizing the zero-shot capability of GPT-3.5. Subsequently, this dataset is utilized for fine-tuning a smaller language model. One major challenge of this method is adapting it to different KG construction scenarios. Especially, the few-shot methods are more resource-efficient than the fine-tuned solutions \cite{wornow2024zero}.

An iterative LLM prompting-based pipeline for automatically generating knowledge graphs, which bypasses the need for predefined sets or external ontologies, was proposed by \cite{carta2023iterative}. This pipeline employs a sequence of well-formed LLM prompts for each stage, enabling the identification of relevant entities, extracting their descriptions and types, and identifying meaningful relationships. The authors proposed an approach to entity/relation resolution using semantic aggregation and LLM prompting. It starts with semantic aggregation, calculating similarity scores for entities and relations based on label similarity, entity type similarity, and description similarity using methods like Levenshtein distance and cosine similarity with the Universal Sentence Encoder model. The entities and relations are aggregated if their scores exceed predefined thresholds. Even though the proposed approaches present several advantages, it has certain limitations: (1) The entity/relation resolution phase aggregates nodes and relations having the same meaning, and then the LLM suggests a representative for each cluster based on the cluster elements. This could hinder the precision of the graph, especially if "bike" and "motorcycle" need to be separated. Still, the model merges them into "vehicle." (2) The latter phase involves post-processing, which could be computationally intensive. (3) The post-processing phase assumes that entities and relations are extracted. Hence, if entities are not resolved before relation extraction, redundant relations from redundant entities could arise, worsening the quality of the relation extraction. 

A comprehensive quantitative and qualitative evaluation of LLMs for KG construction and reasoning was provided \cite{zhu2023llms}, using eight diverse datasets across four representative tasks: entity and relation extraction, event extraction, link prediction, and question-answering. Key findings reveal that while GPT-4 performs well in KG construction tasks, it excels even more in reasoning tasks, sometimes surpassing fine-tuned models. The paper also proposes AutoKG, a multi-agent-based approach that utilizes LLMs and external sources for KG construction and reasoning.

\section{Incremental Text2KG}

This work aims to develop a plug-and-play solution for constructing KGs from documents with resolved entities and relations as output. Adopting a 'zero-shot' approach is essential to ensure the solution's applicability across various KG construction scenarios. This approach means that the prompts used to generate the KG do not require prior examples or predefined ontologies.

\subsection{Problem Formulation}
A graph can be defined as $\mathcal{G} = (\mathcal{E}, \mathcal{R})$ where $\mathcal{E}$ is the set of nodes and $\mathcal{R}$ denotes the set of edges \cite{jin2023large}. Considering the difficulty in merging similar concepts, we defined two constraints for the solution:

\begin{enumerate}[label= (C\arabic*)]
    \item An entity $e_{i} \in \mathcal{E}$, the set of entities and a relation $r_{k} \in \mathcal{R}$, the set of relations, should each describe a semantically unique concept. \label{c1}
    \item The sets of entities and relations should contain semantically unique elements. This means each entity and relation within the knowledge graph must be distinct and unique, with no duplication or semantic overlaps. \label{c2}

\end{enumerate}
These constraints can be mathematically formulated as follows.

\begin{equation}
    \forall e_i, e_j \in \mathcal{E}, \; i \neq j \Rightarrow e_i \neq e_j
\end{equation}
\begin{equation}
    \forall r_k, r_l \in \mathcal{R}, \; l \neq k \Rightarrow r_k \neq r_l
\end{equation}

\subsection{Proposed method}

\begin{figure}[ht]
\includegraphics[width=1\textwidth]{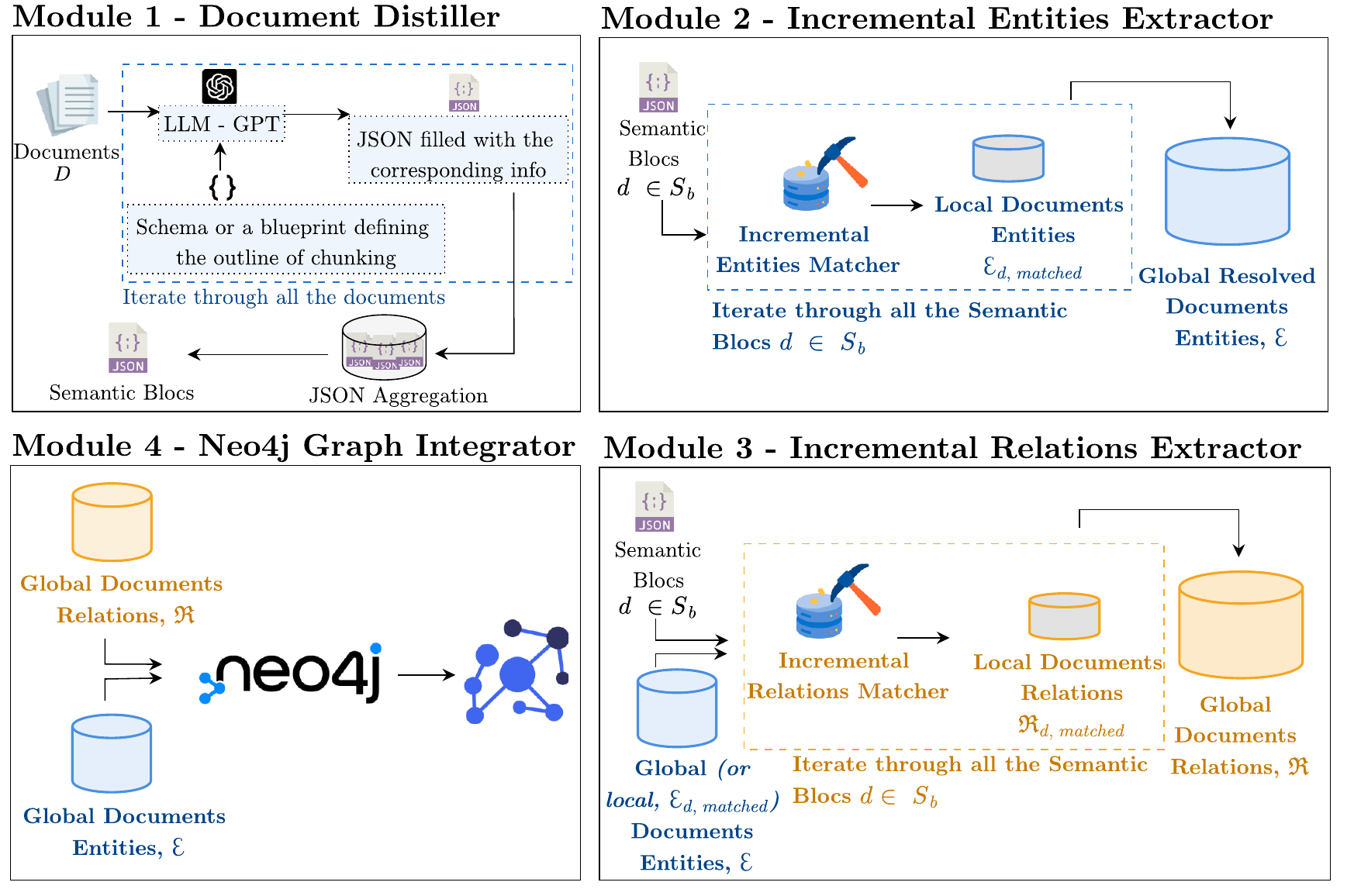}
\centering
\caption{\label{fig:itext2graph} The overall workflow of the \texttt{iText2KG} modules. Module~3, the Incremental Relations Extractor, operates differently depending on whether global or local document entities are provided as context.}
\end{figure}

We propose the \texttt{iText2KG} approach composed of four modules (see Figure~\ref{fig:itext2graph}): Document Distiller, Incremental Entities Extractor, Incremental Relations Extractor, and Neo4j Graph Integrator. 
Each module fulfills a distinct role in constructing the KG. Notably, entity extraction and relation extraction tasks are separated following results described in \cite{carta2023iterative} that positively impact the performance. Further details of modules 1 to 3 are as follows, with the fourth module serving to visualize the graph.

\subsubsection{Module 1 - Document Distiller:}
This module uses LLMs to rewrite input documents into semantic blocks, considering a predefined schema or blueprint. It is important to note that the schema is not an ontology but a blueprint that biases the LLM towards specific classes while maintaining flexibility in others. Practically, the schema functions like a predefined JSON, instructing the LLM to extract particular values (textual information) for specific keys from each document. Some examples of blueprints are available in the \texttt{iText2KG} Github repository. For each document, we will obtain a JSON semi-filled with the desired information if it exists in the document. Then, we aggregate all these semi-filled JSONs to form the semantic blocks of the documents. We have used Langchain's JSON Parser\footnote{\url{https://python.langchain.com/v0.1/docs/modules/model_io/output_parsers/types/json/}}
to define the schema along with the documents as context. The main goals of this module are: (a) To improve the signal-to-noise ratio by reducing noise that may pollute the graph with redundant information. (b) To guide the graph construction process using the schema, especially for concept keys. For example, for a scientific article, we could extract the “title” and the “authors” and add relations like “HAS TITLE" and “HAS AUTHORS” in addition to the semantic information. To ensure the applicability of our solution across various use cases, the schema is an input that depends on user preferences and the particularity of the use case. The idea of reformulating raw documents to enhance the graph construction process has been proven by the following papers \cite{zhang2024attackg+,sun2024knowledge}. The two papers as mentioned earlier introduced a rewriter module, but it depends on the article's specific use case. However, our module is adaptable to many use cases.

\subsubsection{Module 2 – Incremental Entities Extractor:} 
\begin{figure}[ht]
\centering
\includegraphics[width=0.6\textwidth]{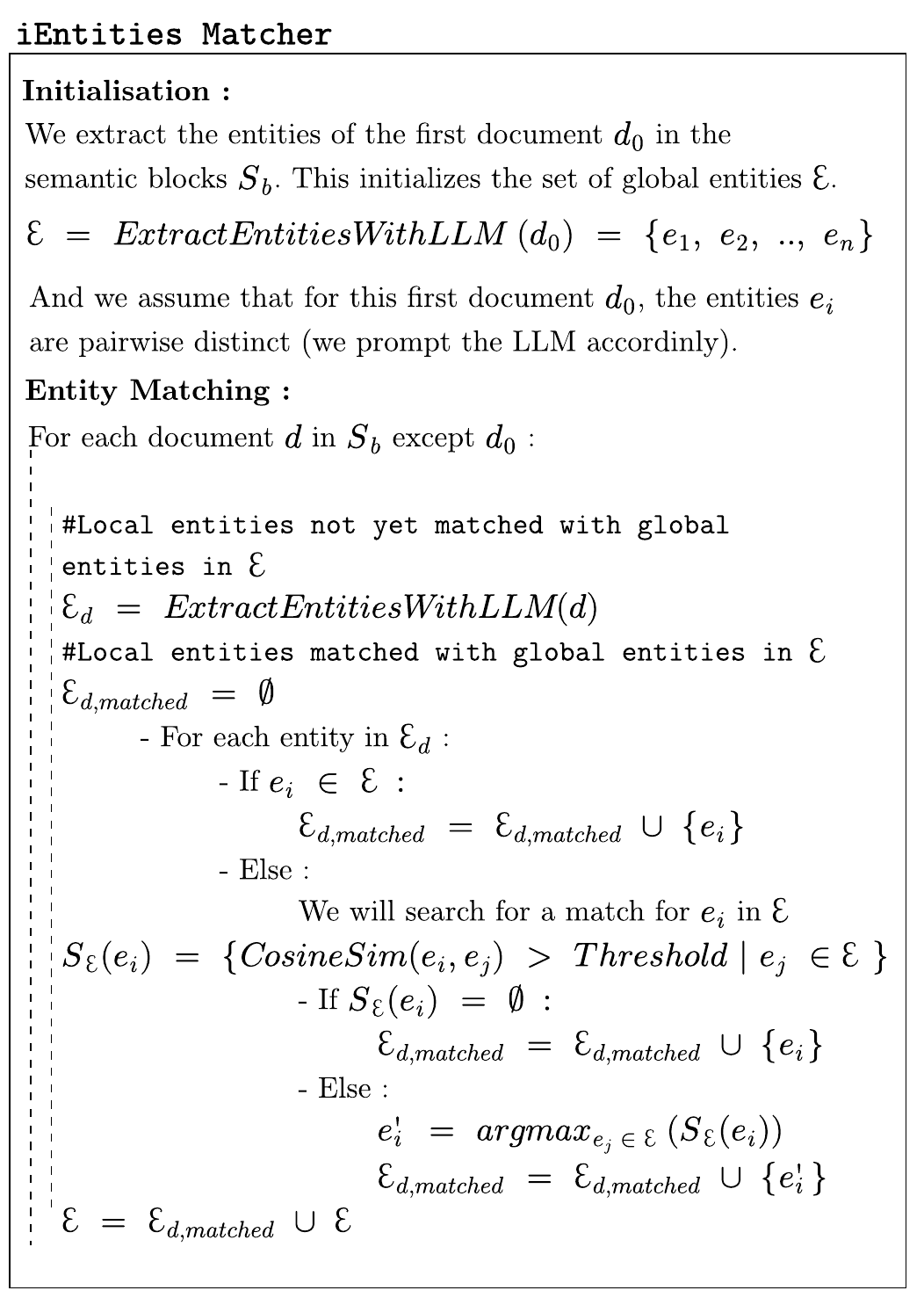}
\caption{\label{fig:ientitiesmatcher}The algorithm of \texttt{iEntities Matcher}}
\end{figure}

The Incremental Entities Matcher (\texttt{iEntities Matcher}) iterates over all the Semantic Blocks and extracts the Global Document Entities. The main algorithm of iEntities Matcher is presented in Figure~\ref{fig:ientitiesmatcher}. Initially, entities are extracted from the first semantic block (document) \( d_0 \) using an LLM, forming the global entity set \( \mathcal{E} \) under the assumption that these entities are pairwise distinct for this first iteration only. 
Considering the constraint \ref{c1}, the LLM is prompted to extract entities representing one unique concept to avoid semantically mixed entities (prompts are presented in the \texttt{iText2KG} GitHub repository). 

For subsequent documents \( d \) in \( D \), the algorithm extracts local entities \( \mathcal{E}_d \). It then attempts to match these local entities with the global entities in \( \mathcal{E} \). If a local entity \( e_i \) is found in \( \mathcal{E} \), it is added to the matched set \( \mathcal{E}_{d,\text{matched}} \). If not, the algorithm searches for a similar entity in \( \mathcal{E} \) using a cosine similarity measure with a predefined threshold. If no match is found, the local entity is added to \( \mathcal{E}_{d,\text{matched}} \); otherwise, the best matching global entity \( e'_i \) (based on maximum similarity) is added. The global entity set \( \mathcal{E} \) is then updated by unifying it with \( \mathcal{E}_{d,\text{matched}} \). This process is repeated for each document in \( D \), resulting in a comprehensive global entity set \( \mathcal{E} \).

\subsubsection{Module 3 – Incremental Relations Extractor:} 
\begin{figure}[ht]
\centering
\includegraphics[width=0.65\textwidth]{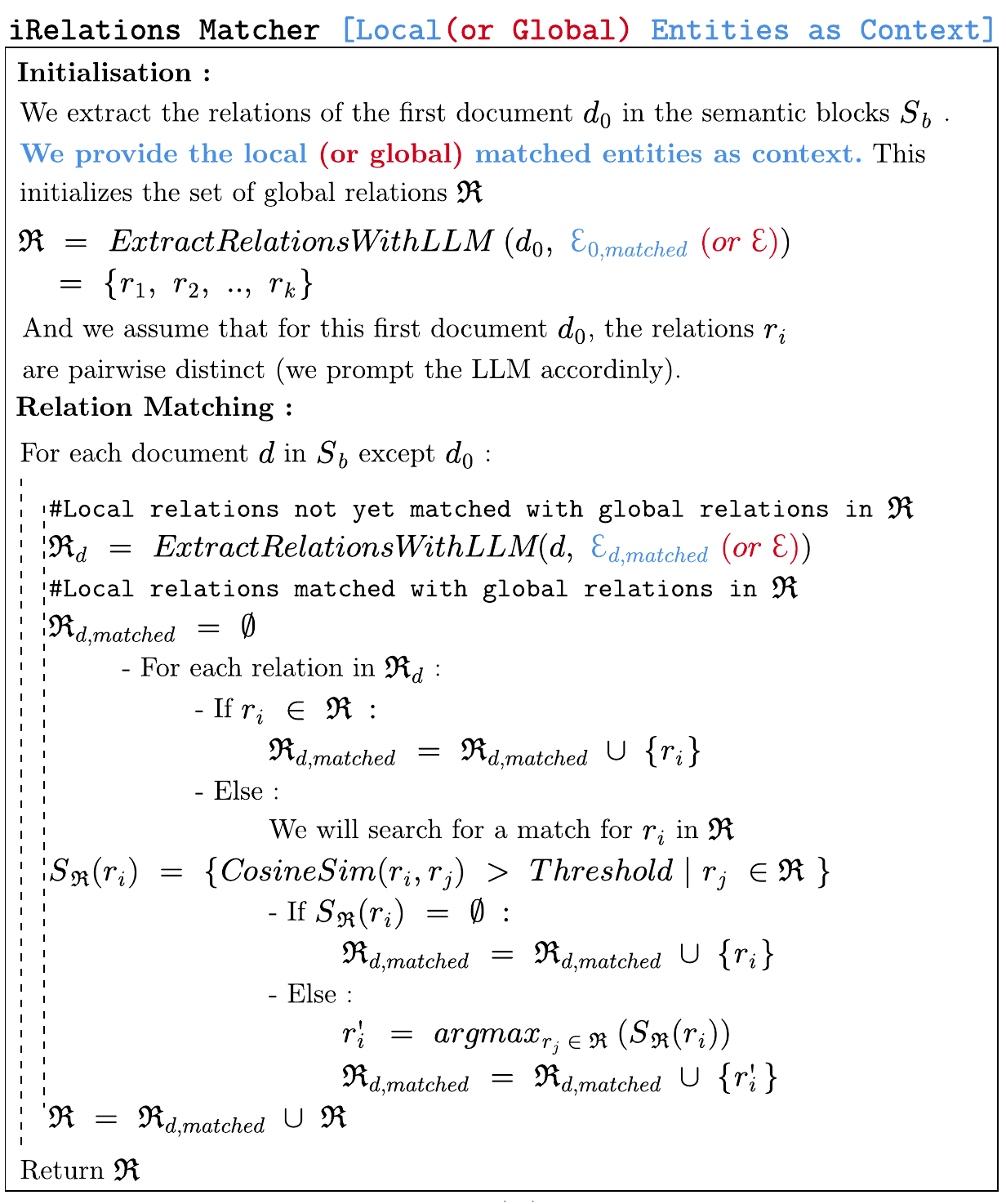}
\caption{\label{fig:irelationsmatcher} The two versions of \texttt{iRelations Matcher}}
\end{figure}

The Global Document Entities $\mathcal{E}$ are provided as context along with each Semantic Block to the Incremental Relations Matcher (\texttt{iRelations Matcher}) to extract the Global Document Relations. The same approach used for \texttt{iEntities Matcher} applies here. We have observed different behaviors in relation extraction depending on whether global or local entities are used as context with the Semantic Block for the LLM. When global entities are provided as context, the LLM extracts both the relations directly stated and implied by the Semantic Block, especially for entities not explicitly present in the Semantic Block. This enriches the graph with potential information but increases the likelihood of irrelevant relations. Conversely, when locally matched entities are provided as context, the LLM only extracts the relations directly stated by the context. This approach reduces the richness of the graph but also lowers the probability of irrelevant relations. The two versions of \texttt{iRelations Matcher} are presented in Figure~\ref{fig:irelationsmatcher}. This result will be further discussed in Section~\ref{experiments}.

\subsubsection{Module 4 – Graph Integrator:} The Global Document Entities and the Global Document Relations are fed into Neo4j to construct the knowledge graph.

\section{Experiments}
\label{experiments}
We chose GPT-4 in all our experiments due to its performance in KG construction and reasoning capabilities, as demonstrated by \cite{zhu2023llms}. Notably, GPT-4 achieves near fine-tuned state-of-the-art performance, even in zero-shot scenarios. To validate our method, it is essential first to evaluate Module~1 to ensure the concordance of the extracted information with the schema and the semantics of the input documents. Moreover, evaluating modules~1 and 2 regarding the extracted triplets and the quality of entity/relation resolution is also important. To ensure the applicability of our method across different KG construction scenarios, we have adopted three use cases: website to KG, scientific article to KG, and Curriculum Vitae to KG.

We have adapted the metrics proposed by \cite{zhang2024attackg+} for Module~1 to our use cases. Hence, we propose the following metrics:

\begin{itemize}
\item \textbf{Schema consistency:} Evaluate whether the content of the rewritten text matches the input schema (the blueprint). For each key presented in the schema, we define $C_{s}(k)$ as the number of elements correctly matched to the schema related to the key $k$. $I_{s}(k)$ as the number of elements that were added but did not belong to the schema. The consistency score for a key in the schema is:

\begin{equation}
\text{SC(k)} = \frac{C_{s}(k) - I_{s}(k)}{T_{s}(k)}
\end{equation}

Such as:

$T_{s}(k)$: The total elements in the schema corresponding to the key $k$.

If $C_{s}(k) < I_{s}(k)$, $SC(k) = 0$.

Hence, the schema consistency score is : 

\begin{equation}
\text{SC} = \sum_{k \in K} \frac{SC(k)}{card(K)}
\end{equation}

Where $K$ is the set of keys of the schema.

\item \textbf{Information consistency:} Evaluate whether the rewritten text's content matches the original report's semantics, categorized as follows: very different (<30\%), medium (30-60\%), largely consistent (60-90\%), and fully consistent (>90\%).
\end{itemize}

For the second and third modules, it is important to ensure that the extracted entities and relations are resolved and that the extracted triplets are relevant to the input documents. Therefore, we propose the following metrics:

\begin{itemize}
\item \textbf{Triplet Extraction Precision:}  Evaluate the consistency of the triplets with the corresponding text regardless of the entity/relation resolution process. It is important to note that a relevant triplet is implied and not necessarily directly stated by the text. We define the precision score as the number of extracted relevant triplets divided by the total number of extracted triplets.

\item \textbf{Entity/Relation Resolution False Discovery Rate:} Evaluate the proportion of unresolved (false positive) entities or relations among the total extracted entities or relations. Specifically, we calculate the ratio of unresolved entities or relations to the total number of extracted entities or relations. This metric provides a clear indication of the reliability of the entity and relation extraction process by highlighting the proportion of errors (unresolved entities/relations) within the total extractions. 
\end{itemize}

\subsection{Datasets and Baseline Methods}
To evaluate \texttt{Document Distiller}, we have generated 5 CVs using GPT-4, selected 5 company websites, and 5 scientific articles. It is important to note that we have extracted the textual information from websites, which will serve as input to our model.

To evaluate the consistency of triplets extracted by \texttt{iEntities Extractor} and \texttt{iRelations Extractor}, we used the annotated dataset from \cite{kabal&al2024}. We observed that this dataset is not exhaustive for triplet extraction, leading us to conduct manual checks for triplets not present in the dataset. This manual check combined with the aforementioned dataset composes the ground truth. To assess the False Discovery Rate of the entity/relation resolution process, we performed the KG construction process using different baseline methods.

We have compared our method against baseline methods including Graph Construction using OpenAI Function Method\footnote{\url{https://github.com/tomasonjo/blogs/blob/master/llm/openaifunction_constructing_graph.ipynb}}
, Langchain\footnote{\url{https://python.langchain.com/v0.1/docs/use_cases/graph/constructing/}}
, and LlamaIndex\footnote{\url{https://docs.llamaindex.ai/en/stable/examples/property_graph/property_graph_basic/}}. 

\subsection{First Module Evaluation Results}

\subsubsection{Schema Consistency}

Table~\ref{table:1} demonstrates that \texttt{Document Distiller} achieves high schema consistency across various document types. Scientific articles and CVs exhibit the highest schema consistency scores, indicating the module's capability to handle structured information, particularly for documents where the data is primarily organized using titles. While still achieving a strong score of 0.94, websites present a slightly lower consistency, which may be attributed to web content's varied and less structured nature. These results highlight the robustness and adaptability of \texttt{Document Distiller} in processing and extracting structured information from diverse document types.

\begin{table}[h!]
\caption{The Schema Consistency Score for the different types of documents.}
\centering
\begin{tabular}{lllll}
\cline{1-4}
\multicolumn{1}{|c|}{\textbf{Documents}}                & \multicolumn{1}{c|}{\textbf{CVs}} & \multicolumn{1}{c|}{\textbf{Scientific Articles}} & \multicolumn{1}{c|}{\textbf{Websites}} &  \\ \cline{1-4}
\multicolumn{1}{|c|}{\textbf{Schema Consistency Score}} & \multicolumn{1}{c|}{0.97 $\pm$ 0.09}         & \multicolumn{1}{c|}{0.98 $\pm$ 0.04}                         & \multicolumn{1}{c|}{0.94 $\pm$ 0.13}              &  \\ \cline{1-4}
\end{tabular}
\label{table:1}
\end{table}
\subsubsection{Information Consistency}
\begin{figure}[h]
\centering
\includegraphics[width=0.8\textwidth]{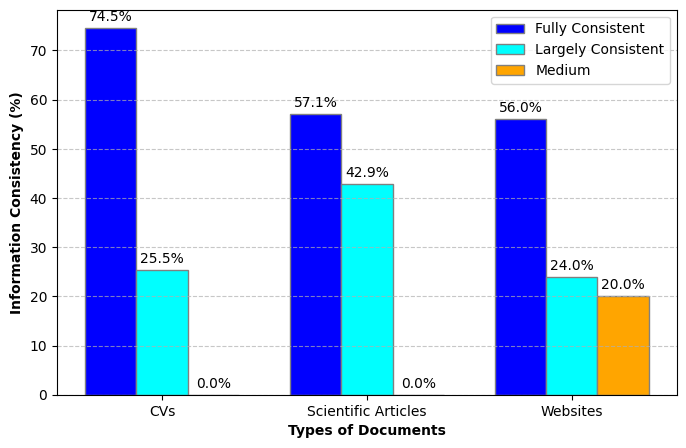}
\caption{\label{fig:info_cons} Bar Plot of the Information Consistency Scores for the different types of Documents}
\end{figure}

Figure~\ref{fig:info_cons} illustrates the information consistency across different types of documents: CVs, scientific articles, and websites. For CVs, the majority of the information (74.5\%) is fully consistent, with 25.5\% being largely consistent and no medium consistency. This indicates that the rewritten text closely matches the semantics of the original content for CVs. This is because CVs are primarily written in clear and concise phrases, making it easier for the LLM to capture the semantics. In the case of scientific articles, 57.1\% of the information is fully consistent, and 42.9\% is largely consistent, showing a high degree of accuracy in preserving the original semantics, though slightly less than CVs. This is predictable, especially since scientific articles are written in scientific English with more complex phrases. Websites have 56.0\% of information fully consistent, 24.0\% largely consistent, and 20.0\% medium consistency. This may be due to the unstructured nature of web content, which poses a greater challenge for accurate semantic rewriting.

\subsection{Second and Third Modules Evaluation Results}

\subsubsection{Triplet Extraction} 

Table~\ref{table:2} shows different behaviors in relation extraction depending on whether global or local entities are used as context with the Semantic Block for the LLM. The precision of relevant triplets when global entities are fed as context is 10\% lower than that of relevant triplets when local entities are fed as context. When global entities are used as context, the LLM extracts relations explicitly mentioned and implied within the Semantic Block. This results in a richer graph with more potential information and a higher chance of irrelevant relations. On the other hand, using locally matched entities as context leads the LLM to extract only the directly stated relations, resulting in a less enriched graph but with a lower likelihood of irrelevant relations.

\begin{table}[ht]
\caption{Precision scores for relevant triplets across two datasets: music and computer science. The scores are presented for Global Entities as Context and Local Entities as Context.}
\centering
\begin{tabular}{cccl}
\cline{1-3}
\multicolumn{1}{|c|}{\textbf{}}                         & \multicolumn{1}{c|}{\textbf{Global Entities}} & \multicolumn{1}{c|}{\textbf{Local Entities}} &  \\ \cline{1-3}
\multicolumn{1}{|c|}{\textbf{Computer Science Dataset}} & \multicolumn{1}{c|}{0.83 $\pm$ 0.06}                                & \multicolumn{1}{c|}{0.94 $\pm$ 0.06}                               &  \\ \cline{1-3}
\multicolumn{1}{|c|}{\textbf{Music Dataset}}            & \multicolumn{1}{c|}{0.81 $\pm$ 0.05}                                & \multicolumn{1}{c|}{0.9 $\pm$ 0.07}                &  \\ \cline{1-3}
\multicolumn{1}{l}{}                                    & \multicolumn{1}{l}{}                                     & \multicolumn{1}{l}{}                                    & 
\end{tabular}
\label{table:2}
\end{table}

This presents a trade-off that depends on the use case. We leave it to the user to decide whether to accept a 10\% decrease in precision in exchange for an enriched graph or to gain 10\% precision with a less enriched graph.

\subsubsection{Entity/Relation Resolution}
To the best of our knowledge, LlamaIndex constructs unconnected sub-graphs with edge-level and node-level textual information for retrieval-augmented generation (RAG); hence, we did not evaluate LlamaIndex against our method. From Table~\ref{table:3} and Table~\ref{table:4}, we conclude that our method delivers superior results for the entity and relation resolution process across three different KG construction scenarios: scientific articles to KG, CVs to KG, and websites to KG. Additionally, the results indicate that when the number of input documents is small and they are structured with clear, non-complex phrases, the LLM performs well in entity and relation resolution, as demonstrated by the CVs to KG process. 

\begin{table}[ht]
\caption{False Discovery Rates of unresolved entities for entity resolution process across three KG construction scenarios. }
\centering
\begin{tabular}{|c|c|c|c|c|}
\hline
\textbf{}                    & \textbf{OpenAI Function} & \textbf{Langchain} & \textbf{LlamaIndex} & \textbf{Our Method} \\ \hline
\textbf{Scientific Articles} & 0.11 $\pm$ 0.04           & 0.14 $\pm$ 0.08               & -                   & 0.01 $\pm$ 0.01                \\ \hline
\textbf{CVs}                 & 0              & 0                  & -                   & 0                   \\ \hline
\textbf{Websites}            & 0.31 $\pm$ 0.05           & 0.29 $\pm$ 0.06               & -                   & 0                   \\ \hline
\end{tabular}
\label{table:3}
\end{table}

\begin{table}[ht]
\caption{False Discovery Rates of unresolved relations for relation resolution process across three KG construction scenarios.}
\centering
\begin{tabular}{|c|c|c|c|c|}
\hline
\textbf{}                    & \textbf{OpenAI Function} & \textbf{Langchain} & \textbf{LlamaIndex} & \textbf{Our Method} \\ \hline
\textbf{Scientific Articles} & 0.07 $\pm$ 0.01          & 0.06 $\pm$ 0.01               & -                   & 0.01 $\pm$ 0.01               \\ \hline
\textbf{CVs}                 & 0              & 0                  & -                   & 0                   \\ \hline
\textbf{Websites}            & 0.15 $\pm$ 0.01          & 0.14 $\pm$ 0.02              & -                   & 0                   \\ \hline
\end{tabular}
\label{table:4}
\end{table}

Moreover, the False Discovery Rates of unresolved entities and relations for websites to KG are higher than in the other KG construction scenarios. This is due to the larger number of documents (chunks) and the unstructured nature of website textual information. Consequently, without an effective resolution process, the LLM struggles to map similar entities or relations. Therefore, as long as the number of documents (chunks) is large and the text is unstructured with complex language, the entity/relation resolution process becomes crucial for building consistent KGs.

\subsubsection{Threshold Estimation}
To estimate the threshold for merging entities and relationships based on cosine similarity, a dataset of 1,500 similar entity pairs and 500 relationships, inspired by various domains (e.g., news, scientific articles, HR practices), was generated using GPT-4 and is available in the \texttt{iText2KG} GitHub repository. Entities and relationships were vectorized using the pre-trained model \texttt{text-embedding-3-large}\footnote{\url{https://platform.openai.com/docs/guides/embeddings/embedding-models}}. The mean and standard deviation of cosine similarity for these datasets were then calculated (Table \ref{table:5}). An upper threshold (e.g., 0.7) was chosen to ensure high precision, while a lower threshold reduced resolution specificity.

\begin{table}[h!]
\caption{Cosine Similarities of the Two Datasets for Entity and Relationship Resolution.}
\centering
\begin{tabular}{ll}
\cline{1-2}
\multicolumn{1}{|c|}{\textbf{Entities Dataset}}       & \multicolumn{1}{c|}{\textbf{Relationships Dataset}} \\ \cline{1-2}
\multicolumn{1}{|c|}{0.6 $\pm$ 0.12}                 & \multicolumn{1}{c|}{0.56 $\pm$ 0.1}                \\ \cline{1-2}

\end{tabular}
\label{table:5}
\end{table}
To illustrate the results of KG construction, Figure~\ref{fig:text_to_kg} presents a comparison between baseline methods and \texttt{iText2KG} across three distinct scenarios. The observations are as follows:
\begin{figure}[ht]
\centering
\includegraphics[width=0.9\textwidth]{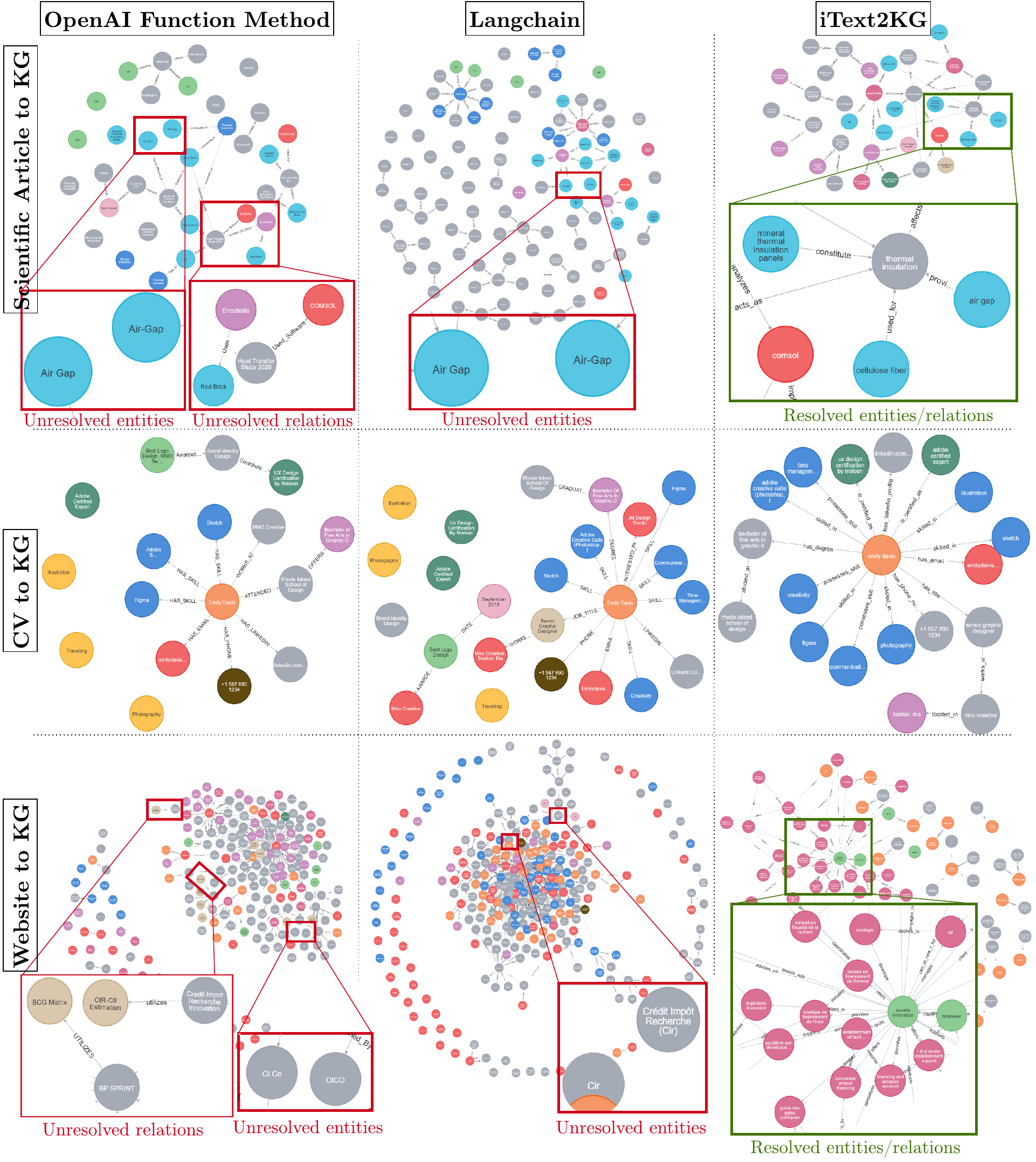}
\caption{\label{fig:text_to_kg} Comparison of KG construction across three scenarios between baseline methods and our method, \texttt{iText2KG}.}
\end{figure}

\begin{itemize}
    \item The baseline methods reveal the presence of isolated nodes without relations in all three KG construction scenarios. This phenomenon may be attributed to the simultaneous execution of entity extraction and relations extraction, which can induce hallucinatory effects in language models, leading to a "forgetting" effect. This observation supports the findings of \cite{carta2023iterative}, which suggest that separating the processes of entity and relation extraction can enhance performance.

    \item From the 'Website to KG' scenario, an increase in the volume of input documents is associated with the emergence of noisy nodes within the graph. This underscores the critical need for Module~1 to effectively refine and distill the input data.

    \item The \texttt{iText2KG} method demonstrates improved entity and relation resolution across the three KG construction scenarios. According to the data in Table~\ref{table:3} and Table~\ref{table:4}, when input documents are fewer and composed of straightforward, non-complex phrases, the language model shows high efficiency in entity and relation resolution, as evidenced in the 'CVs to KG' process. Conversely, the challenges increase with more complex and voluminous data sets, as shown in the 'Website to KG' scenario.

\end{itemize}

Moreover, it is important to highlight the effect of the chunking size of the input document and the threshold on KG construction. Input documents to the Document Distiller can be independent documents or chunks. If the chunk size is smaller, the semantic blocks will capture more specific details from the documents, and vice versa.

\section{Conclusion}
In this paper, we introduced \texttt{iText2KG}, an approach for incremental KG construction leveraging the zero-shot capabilities of LLMs. Our methodology addressed limitations inherent in traditional KG construction processes, which typically depend on predefined ontologies and extensive supervised training. 

A key advantage of the \texttt{iText2KG} approach is its flexibility, which stems from the use of a user-defined blueprint that outlines the key components to extract during KG construction. This allows the method to adapt to a wide range of scenarios, as there is no universal blueprint for all use cases; instead, the design varies depending on the specific application. Moreover, The \texttt{iText2KG} method achieves document-type independence by using a flexible, user-defined blueprint to guide the extraction process, allowing it to handle both structured and unstructured texts.

Empirical evaluations across diverse contexts, such as scientific documents, web content, and CVs, demonstrated the superior performance of the \texttt{iText2KG} approach compared to established baseline methods. The method achieves enhanced schema consistency and high precision in entity and relation extraction, effectively mitigating issues related to semantic duplication and unresolved entities, which are prevalent in traditional methodologies.

Future research will focus on enhancing metrics such as cosine similarity for advanced entity and relation matching, eliminating the necessity to define a threshold as a hyperparameter, and integrating the entity type as a parameter of the matching process.


%
%
%
\bibliographystyle{splncs04}
%
\bibliography{sample}

\begin{thebibliography}{10}
\providecommand{\url}[1]{\texttt{#1}}
\providecommand{\urlprefix}{URL }
\providecommand{\doi}[1]{https://doi.org/#1}

\bibitem{carta2023iterative}
Carta, S., Giuliani, A., Piano, L., Podda, A.S., Pompianu, L., Tiddia, S.G.: Iterative zero-shot {LLM} prompting for knowledge graph construction. arXiv preprint arXiv:2307.01128  (2023)

\bibitem{ding2024automated}
Ding, L., Zhou, S., Xiao, J., Han, J.: Automated construction of theme-specific knowledge graphs. arXiv preprint arXiv:2404.19146  (2024)

\bibitem{eberendu2016unstructured}
Eberendu, A.C., et~al.: Unstructured data: an overview of the data of big data. International Journal of Computer Trends and Technology  \textbf{38}(1),  46--50 (2016)

\bibitem{hu2023llm}
Hu, Y., Zou, F., Han, J., Sun, X., Wang, Y.: {LLM-Tikg}: Threat intelligence knowledge graph construction utilizing large language model. Available at SSRN 4671345  (2023)

\bibitem{jin2023large}
Jin, B., Liu, G., Han, C., Jiang, M., Ji, H., Han, J.: Large language models on graphs: A comprehensive survey. arXiv preprint arXiv:2312.02783  (2023)

\bibitem{kabal&al2024}
Kabal, O., Harazallah, M., Guillet, F., Ichise, R.: Enhancing domain-independent knowledge graph construction through {OpenIE} cleaning and llms validation {(G-T2KG)}. In: 28th International Conference on Knowledge-Based and Intelligent Information \& Engineering Systems (KES 2024) (2024), to appear

\bibitem{kommineni2024human}
Kommineni, V.K., K{\"o}nig-Ries, B., Samuel, S.: From human experts to machines: An {LLM} supported approach to ontology and knowledge graph construction. arXiv preprint arXiv:2403.08345  (2024)

\bibitem{mihindukulasooriya2023text2kgbench}
Mihindukulasooriya, N., Tiwari, S., Enguix, C.F., Lata, K.: Text2kgbench: A benchmark for ontology-driven knowledge graph generation from text. In: International Semantic Web Conference. pp. 247--265. Springer (2023)

\bibitem{nasar2021named}
Nasar, Z., Jaffry, S.W., Malik, M.K.: Named entity recognition and relation extraction: State-of-the-art. ACM Computing Surveys (CSUR)  \textbf{54}(1),  1--39 (2021)

\bibitem{singh2018natural}
Singh, S.: Natural language processing for information extraction. arXiv preprint arXiv:1807.02383  (2018)

\bibitem{sun2024knowledge}
Sun, Z., Ting, Y.S., Liang, Y., Duan, N., Huang, S., Cai, Z.: Knowledge graph in astronomical research with large language models: Quantifying driving forces in interdisciplinary scientific discovery. arXiv preprint arXiv:2406.01391  (2024)

\bibitem{wornow2024zero}
Wornow, M., Lozano, A., Dash, D., Jindal, J., Mahaffey, K.W., Shah, N.H.: Zero-shot clinical trial patient matching with {LLMs}. arXiv preprint arXiv:2402.05125  (2024)

\bibitem{zhang2024attackg+}
Zhang, Y., Du, T., Ma, Y., Wang, X., Xie, Y., Yang, G., Lu, Y., Chang, E.C.: {AttacKG+}: Boosting attack knowledge graph construction with large language models. arXiv preprint arXiv:2405.04753  (2024)

\bibitem{zhu2023llms}
Zhu, Y., Wang, X., Chen, J., Qiao, S., Ou, Y., Yao, Y., Deng, S., Chen, H., Zhang, N.: {LLMs} for knowledge graph construction and reasoning: Recent capabilities and future opportunities. arXiv preprint arXiv:2305.13168  (2023)

\end{thebibliography}

\end{document}